\documentclass[runningheads]{llncs}
\usepackage{graphicx}
\usepackage{wrapfig}
\usepackage{comment}
\usepackage{color}
\usepackage{epsfig}
\usepackage{booktabs}       
\usepackage{mathtools}
\usepackage{setspace}
\usepackage{algorithm}
\usepackage[noend]{algpseudocode}
\usepackage{lipsum}
\usepackage{amssymb,amsmath,amsfonts}
\usepackage{caption}
\usepackage{overpic}
\usepackage[colorlinks=true]{hyperref}

\usepackage[dvipsnames]{xcolor}

\newcommand{\JC}[1]{}
\newcommand{\RR}[1]{}
\newcommand{\ric}[1]{}

\newcommand{\eg}{\textit{e.g.}}
\newcommand{\ie}{\textit{i.e.}}

\begin{document}
\pagestyle{headings}
\mainmatter
\def\ECCVSubNumber{4140}  

\title{Learning Unbiased Representations \\ via Mutual Information Backpropagation}

\titlerunning{Learning Unbiased Representations}
%
\author{Ruggero Ragonesi\inst{1,2} \and
Riccardo Volpi\inst{3}\thanks{Work done while author was working at Istituto Italiano di Tecnologia.} \and
Jacopo Cavazza\inst{1} \and \\
Vittorio Murino\inst{1,4,5}}
\authorrunning{Ragonesi et al.}
%
\institute{PAVIS Department, Istituto Italiano di Tecnologia, Genova, Italy \and
DITEN, University of Genova, Italy \and
Naver Labs Europe, Grenoble, France \and
Department of Computer Science, University of Verona, Italy \and
Huawei Technologies Ltd., Ireland Research Center, Dublin, Ireland \\
\email{\{name.lastname\}@iit.it}}
\maketitle

\begin{abstract}
We are interested in learning data-driven representations that can generalize well, even when trained on inherently biased data. In particular, we face the case where some attributes (bias) of the data, if learned by the model, can severely compromise its generalization properties. We tackle this problem through the lens of information theory, leveraging recent findings for a differentiable estimation of mutual information. We propose a novel end-to-end optimization strategy, which simultaneously estimates and minimizes the mutual information between the learned representation and the data attributes. When applied on standard benchmarks, our model shows comparable or superior classification performance with respect to state-of-the-art approaches. Moreover, our method is general enough to be applicable to the problem of ``algorithmic fairness'', with competitive results.\\

Code publicly available at \url{https://github.com/rugrag/learn-unbiased}
\end{abstract}
\section{Introduction}\label{sec:intro}

The need for proper data representations is ubiquitous in machine learning and computer vision~\cite{Bengio2013RLR}. Indeed, given a learning task, the competitiveness of the proposed models crucially depends upon the data representation one relies on. 
In the last decade, the mainstream strategy for designing feature representations switched from hand-crafting to learning them in a data-driven fashion \cite{collobert2011natural,Alex,szegedy2015going,mnih2015human,he2016deep,huang2017densely,xie2018aggregated}. In this context, deep neural networks have shown an extraordinary efficacy in learning hierarchical representations via backpropagation~\cite{BackProp}. However, while learning representations from data allows achieving remarkable results in a broad plethora of tasks, it leads to the following shortcoming: a representation may inherit the intrinsic bias of the dataset used for training. 

\begin{wrapfigure}[19]{r}{0.55\textwidth}
    \centering
    \includegraphics[width=\linewidth]{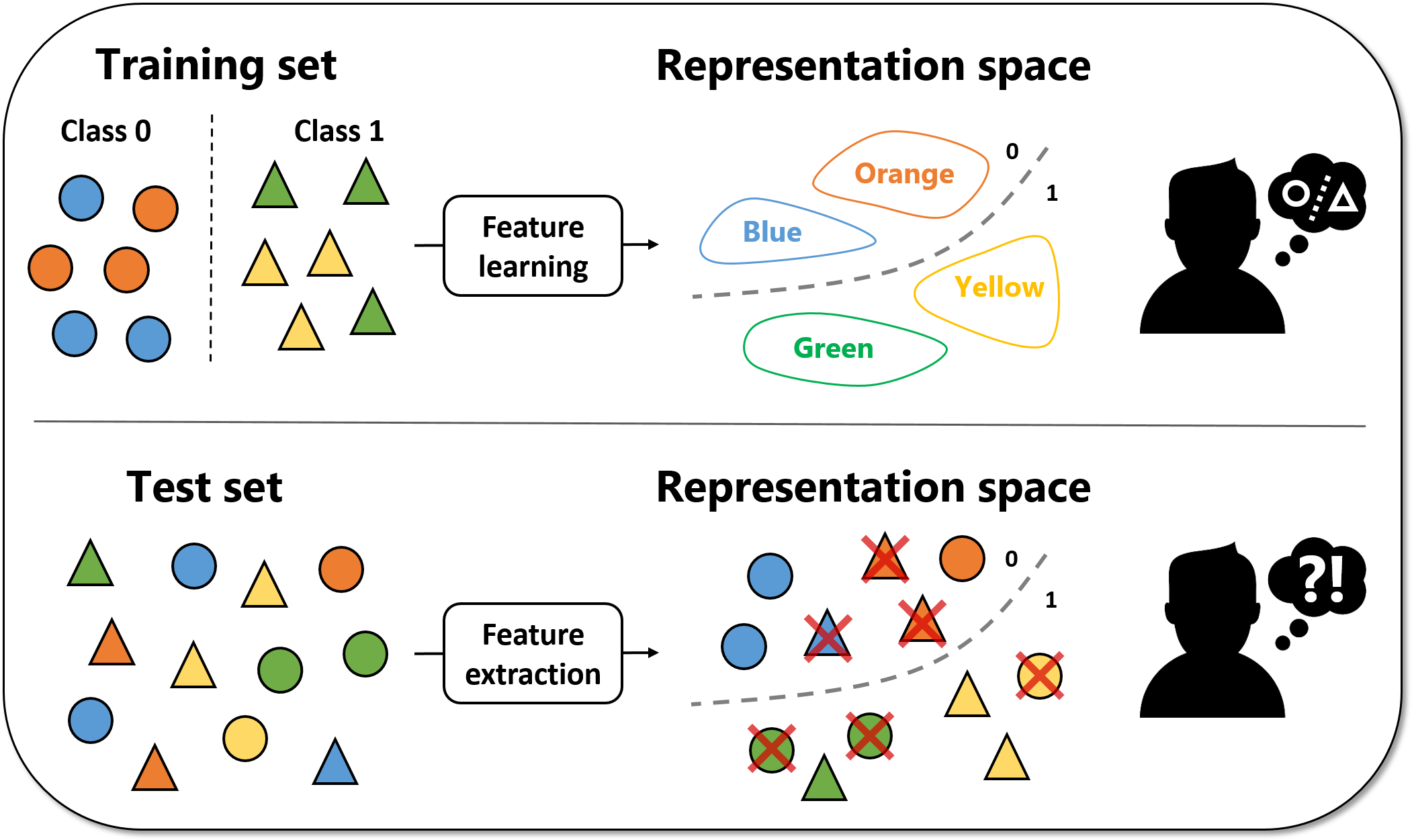}
  \caption{\footnotesize \textbf{Problem setting.} When learning a feature representation from the data itself \textit{(top)}, we may undesirably capture the inherent bias of the dataset (here, exemplified by colors), as opposed to learning the desired patterns (here, represented by shapes). This results in models that poorly generalize when deployed into unbiased scenarios \textit{(bottom)}.}
  \label{fig:applicative_scenario}
\end{wrapfigure}

This is highly undesirable, because it leads a model to poorly generalize in scenarios different from the training one (the so-called ``domain shift'' issue~\cite{NameTheDataset}).

In this paper, we are interested in learning representations that are discriminative for the supervised learning task of interest, while being invariant to certain specified \emph{biased attributes} of the data. By ``biased attribute'', we mean an inherent bias of the dataset, which is assumed to be known and follows a certain distribution during training. At test time, the distribution of such attribute may abruptly change, thus tampering the generalization capability of the model and affecting its performance for the given task~\cite{zisserman2018,moyer2018neurips,kim2019cvpr}. 

One intuitive example is provided in Figure~\ref{fig:applicative_scenario}: we seek to train a \textit{shape classifier}, but each shape has a distinct color -- the biased attribute. Unfortunately, a model can fit the training distribution by discriminating either the color or the shape. Among the two options, we are interested in the latter only, because the first one does not allow generalizing to shapes with different colors. Thus, if we were capable of learning a classifier while unlearning the color, we posit that it would better generalize to shapes with arbitrary colors. 
Like other prior works ~\cite{VFAE_2016_fairness,moyer2018neurips,kim2019cvpr,zisserman2018}, we operate in a scenario where the labels of biased attributes are assumed to be known. An example of application domain in which the hypothesis of having known labels for the bias holds, is algorithmic fairness \cite{kleinberg2016inherent,donini2018empirical,zhang_2018_fairness,wang_2019_fairness}, where the user specifies which attributes the algorithm has to be invariant to (\eg, learning a face recognition system which is not affected by gender or ethnicity biases).


In this paper, we tackle this problem through the lens of information theory. Since mutual information can be used to quantify the nonlinear dependency of the learned feature space with respect to the dataset bias, we argue that a good strategy to face the aforementioned problem is minimizing the mutual information between the learned representation and the biased attributes. This would result in a data representation that is statistically independent from the specified bias, and that, in turn, would generalize better. 


Unfortunately, the estimation of the mutual information is not a trivial problem \cite{poole2019variational}. 
In the context of representation learning, two bodies of work proposed solutions to the problem of learning unbiased representations via information theoretic measures: one that relies on adversarial training ~\cite{zisserman2018,kim2019cvpr}, and one based on variational inference~\cite{moyer2018neurips}. Adversarial methods~\cite{zisserman2018,kim2019cvpr} learn unbiased representations by ``fooling'' a classifier trained to predict the attribute from the learned representation. Such condition is argued to be a proxy for the minimization of the mutual information \cite{kim2019cvpr}. However, since the mathematical principles that govern adversarial training are nowadays still elusive \cite{jin2019local,beyondNash}, a key difficulty is how to properly balance between learning the task and unlearning the attribute. A better control on this aspect can be achieved by the sound theoretical framework of variational inference, which properly formalizes the prior and the conditional dependences among variables. However, when implementing those methods in practice, approximations need to be done to replace the computationally intractable posterior with an auxiliary distribution, but at the cost of several assumptions of independence among the variables. Moreover, such methods are more problematic to scale to complex computer vision tasks, and have been applied mostly on synthetic or toy datasets \cite{VFAE_2016_fairness,moyer2018neurips}.

Due to the aforementioned difficulties, in this paper, we seek to leverage the mathematical soundness of mutual information as a means to avoid adversarial training. To this end, we devise a computational pipeline that relies on a neural estimator for the mutual information (MINE~\cite{belghazi18a}). This module provides a more reliable estimate of the mutual information~\cite{poole2019variational}, while still being fully differentiable and, therefore, trainable via backpropagation~\cite{BackProp}. 
Endowed with this model, we propose a training scheme where we alternate between (i) optimizing the estimator and (ii) learning a representation that is both discriminative for the desired task and statistically independent from the specified bias. 
In practice, first, we train a classifier to minimize the discriminative loss for the given task, regularized by the mutual information between the feature representation and the attributes. Second, we update the MINE parameters in order to tailor the mutual information to the current learned representation. 

A key and strong aspect of the proposed approach is that -- in contrast with adversarial methods -- the module that estimates the mutual information is not competing with the feature extractor. For this reason, MINE can be trained until convergence at every training step, avoiding the need to carefully balance between steps (i) and (ii), and guaranteeing an updated estimate of the mutual information throughout the training process.
In adversarial methods such as~\cite{kim2019cvpr}, where the estimate for the mutual information is modeled via a discriminator that the feature extractor seeks to fool~\cite{Ganin,Ganin2}, one cannot train an optimal discriminator at every training iteration. Indeed, if one trains an optimal bias discriminator, the feature extractor will no longer be able to fool it, due to the fact that gradients will become too small~\cite{arjovsky2017iclr} -- and the adversarial game will not reach optimality. This difference is a key novelty of the proposed computational pipeline, which scores favorably 
with respect to prior work on different computer vision benchmarks, from color-biased classification to age-invariant 
recognition of people attributes. 


Furthermore, a critical aspect of this line of work~\cite{zisserman2018,kim2019cvpr} is how to balance between learning the desired task and ``unlearning'' the dataset bias, which is a core, open issue~\cite{zhang_2018_fairness}. 
The training strategy proposed in this paper allows for a very simple strategy to govern this important problem. 
Indeed, as we will show later in the experimental analysis, a very effective approach is selecting the models whose learned representation distribution has the lowest mutual information with that of the biased attribute. We empirically show that these models are also the ones that better generalize to unbiased settings. Most notably, this also provides us with a simple cross-validation strategy for the crucial hyper-parameters: without using any \textit{validation data}, we can select the optimal model as the one that achieves the best fitting to the data, while better minimizing the mutual information. The importance of this contribution is that, when dealing with biased datasets, also the validation set will likely suffer from the same bias, making hyper-parameter selection a thorny problem. Our proposed method properly responds to this problem, whereas former works have not addressed the issue~\cite{kim2019cvpr}.
\paragraph{Paper outline.} In  Section \ref{sec:rel_work}, we discuss the related literature. 
In Sections \ref{sec:prob_form} and \ref{sec:method}, we formalize the problem and describe the proposed method, which is empirically validated in Section \ref{sec:exp}. Concluding remarks are drawn in Section \ref{sec:conclusion}.

\vspace{-5pt}
\section{Related Work}\label{sec:rel_work}
\vspace{-5pt}


The problem of learning unbiased representations has been explored in several sub-fields. In the following section, we cover the most related literature, with particular focus on works that approach our same problem formulation, highlighting similarities and differences.

In domain adaptation~\cite{Daume2006,Blitzer2006,Saenko2010}, the goal is learning representations that generalize well to a (target) domain of interest, for which only unlabeled -- or partially labeled -- samples are available at training time, leveraging annotations from a different (source) distribution. In domain generalization, the goal is to better generalize to unseen domains, by relying on one or more source distributions~\cite{muandet2013icml,li2017iccv}. Adversarial approaches for domain adaptation \cite{Ganin,Ganin2,ADDA,volpi2018cvpr} and domain generalization~\cite{shankar2018iclr,Zunino2019} are very related to our work: their goal is indeed learning representations that do not contain the domain bias, and therefore better generalize in out-of-distribution settings. 
Differently, in our problem formulation we aim at learning representations that are invariant towards specific attributes that are given at training time. 

A similar formulation is related to the so-called ``algorithmic fairness''~\cite{kleinberg2016inherent}. The problem here is learning representations that do not rely on
sensitive attributes (such as, \textit{e.g.}, gender, age or ethnicity), in order to prevent
from learning discriminant capabilities
towards such protected categories. Our methods can be applied in this setting, in order to minimize the mutual information between the learned representation and the sensitive attribute (interpreted as a bias). In these settings, it is important to notice that a ``fairer'' representation does not necessarily  generalize better than a standard one: the trade-off between accuracy and fairness is termed ``fairness price'' \cite{kleinberg2016inherent,donini2018empirical,zhang_2018_fairness,wang_2019_fairness}.


There is a number of works that share our same goal and problem formulation. Alvi et al.~\cite{zisserman2018} learn unbiased representations through the minimization of a confusion loss, learning a representation that does not inherit information related to specified attributes. Kim et al.~\cite{kim2019cvpr}, similar to us, propose to minimize the mutual information between learned features and the bias. However, they face the optimization problem through adversarial training: in practice, in their implementation~\cite{kim2019cvpr-code}, the authors rely on a discriminator trained to detect the bias as an estimator for the mutual information, and learn unbiased representations by trying to fool this module, drawing inspiration from the solution proposed by Ganin and Lempitsky~\cite{Ganin} for domain adaptation. Moyer et al.~\cite{moyer2018neurips} also introduce a penalty term based on mutual information, to achieve representations that are invariant to some factors. In contrast with related works~\cite{zisserman2018,kim2019cvpr,moyer2018neurips}, it shows that adversarial training is not necessary to minimize such objective, and the problem is approached in terms of variational inference, relying on Variational Auto-Encoders (VAEs~\cite{VAE}). Closely related to Moyer et al., other works~\cite{VFAE_2016_fairness,Zemel_2019_fairness} impose a prior on the representation and the underlying data generative factors (\eg, feature vectors are distributed as a factorized Gaussian).

Our proposed solution does not fit under the class of adversarial approaches~\cite{zisserman2018,kim2019cvpr}, nor it is based on VAE~\cite{moyer2018neurips}, and provides several advantages over both. With respect to adversarial strategies, our method has the advantage of relying on a module estimating the mutual information~\cite{belghazi18a} that is not competing with the network trained to learn an unbiased representation. In our computational pipeline, we do not learn unbiased representation by ``fooling'' the estimator, but by minimizing the information that it measures. The difference is subtle, but brings a crucial advantage: in adversarial methods, the discriminator (estimator) cannot be trained until convergence at every training step, otherwise gradients flowing through it would be close to zero almost everywhere in the 
parameter space~\cite{arjovsky2017iclr}, 
preventing from learning an unbiased representation. 
In our case, the estimator can be trained until convergence at every training step, improving the quality 
of its measure without any drawbacks.
Furthermore, our solution can easily scale to large architectures (\eg, for complex computer vision tasks) in a straightforward fashion. While this is true also for adversarial methods~\cite{zisserman2018,kim2019cvpr}, we posit that it might not be the case for methods based on VAEs~\cite{moyer2018neurips}, where one has to simultaneously train a feature extractor/encoder and a decoder.\\



%

\vspace{-20pt}
\section{Problem Formulation}\label{sec:prob_form}


We operate in a setting where data are shaped as triplets $(\mathbf{x},\mathbf{y},\mathbf{c})$, where $\mathbf{x}$ represents a generic datapoint, $\mathbf{y}$ denotes the ground truth label related to a task of interest and $\mathbf{c}$ encodes a vector of given attributes. We are interested in learning a representation $\mathbf{z}$ of $\mathbf{x}$ that allows performing well on the given task, with the constraint of not retaining information related to $\mathbf{c}$. In other words, we desire to learn a model that, when fed with $\mathbf{x}$, produces a representation $\mathbf{z}$ which is maximally discriminative with respect to $\mathbf{y}$, while being invariant with respect to $\mathbf{c}$. 

In this work, we formalize the invariance of $\mathbf{z}$ with respect to $\mathbf{c}$ through the lens of information theory, imposing a null mutual information $I$. Specifically, we constrain the discriminative training (finalized to learn the task of interest) by imposing $I(Z,C)=0$, where $Z$ and $C$ are the random variables associated with $\mathbf{z}$ and $\mathbf{c}$, respectively. In formul\ae, we obtain the following constrained optimization
\vspace{-2pt}
\begin{align}\label{eq:obj}
\min_{\theta,\psi}~\mathcal{L}_{task}(\theta, \psi), ~~~~ s.t. \phantom{a} ~ I(Z,C)=0
\end{align}
where $\theta$ and $\psi$ define the two sets of parameters of the objective $\mathcal{L}_{task}$, which can be tailored to learn the task of interest. With $\theta$, we refer to the trainable parameters of a module $g_{\theta}$ that maps a datapoint $\mathbf{x}$ into the corresponding feature representation $\mathbf{z}$ (that is, $\mathbf{z} = g_{\theta}(\mathbf{x})$). With $\psi$, we denote the trainable parameters of a classifier that predicts $\tilde{\mathbf{y}}$ from a feature vector $\mathbf{z}$  (that is, $\tilde{\mathbf{y}} = f_{\psi}(\mathbf{z})$). The constraint $I(Z,C) = 0$ does not depend upon $\psi$, but only upon $\theta$, since $\mathbf{z}$ obeys to $p_Z$ and $\mathbf{z} = g_\theta(\mathbf{x})$. 


In order to optimize the objective in~\eqref{eq:obj}, we must adopt an estimator of the mutual information. Before detailing our approach, in the following paragraph we cover the background required for a basic understanding of mutual information estimation, with focus on the path we pursue in this work.\\

\textbf{Background on information theory.} The mutual information between two random variables $X,Z$ is given by
\begin{equation}\nonumber 
I(X,Z) =  \int p_{X,Z}(x,z) \log \dfrac{p_{X,Z}(x,z)}{p_X(x) \cdot p_Z(z)} dxdz,
\end{equation}
where 
$p_{X,Z}$ denotes the joint probability of the two variables and $p_X,p_Z$ represent the two marginals. As an alternative to covariance and other linear indicators of statistical dependence, mutual information can account for generic inter-relationships between $X,Z$, going beyond simple correlation \cite{cavazza2016kernelized,CAVAZZA201925}.

The main drawback with mutual information relates to its difficult computation, since the probability distributions $p_X$, $p_Z$ and $p_{X,Z}$ are not known in practice. 
Recently, a general purpose and efficient estimator for mutual information has been proposed by Belghazi et al.~\cite{belghazi18a}. They propose a neural network based approximation to compute the following lower bound for the mutual information $I$:
\vspace{-2pt}
\begin{align}\label{eq:MINE-approx}
I(X,Z) \geq \widehat{I}_\phi(X,Z) := \max_\phi \left( \mathbb{E}_{p_{X,Z}} [T_\phi] - \log \mathbb{E}_{p_X \cdot p_Z} [\exp T_\phi] \right). 
\end{align}
When implementing $T_\phi$ as a feed-forward neural network, the maximization in Eq.~\eqref{eq:MINE-approx} can be efficiently solved via backpropagation~\cite{belghazi18a}. As a result, we can approximate $I(X,Z)$ with $\widehat{I}_\phi(X,Z)$, the so-called ``Mutual Information Neural Estimator'' (MINE~\cite{belghazi18a}). An appealing aspect of MINE is its fully differentiable nature, that enables end-to-end optimization of objectives that rely on mutual information computations.\\

Endowed with all relevant background, in the following section we detail our approach, which is based on the optimization of a Lagrangian for the objective~\eqref{eq:obj}. By relying on MINE~\cite{belghazi18a}, we can efficiently estimate the mutual information and backpropagate through the different modules, in order to unbias the feature representation which is learnt to solve a given supervised learning task.

\section{Method}\label{sec:method}

In the following, we detail how we approach Eq.~\eqref{eq:obj}, both in terms of theoretical foundations and practical implementation.

\vspace{-10pt}
\subsection{Optimization problem}

\begin{figure*}[!t]
    \centering
    \includegraphics[width=0.9\textwidth]{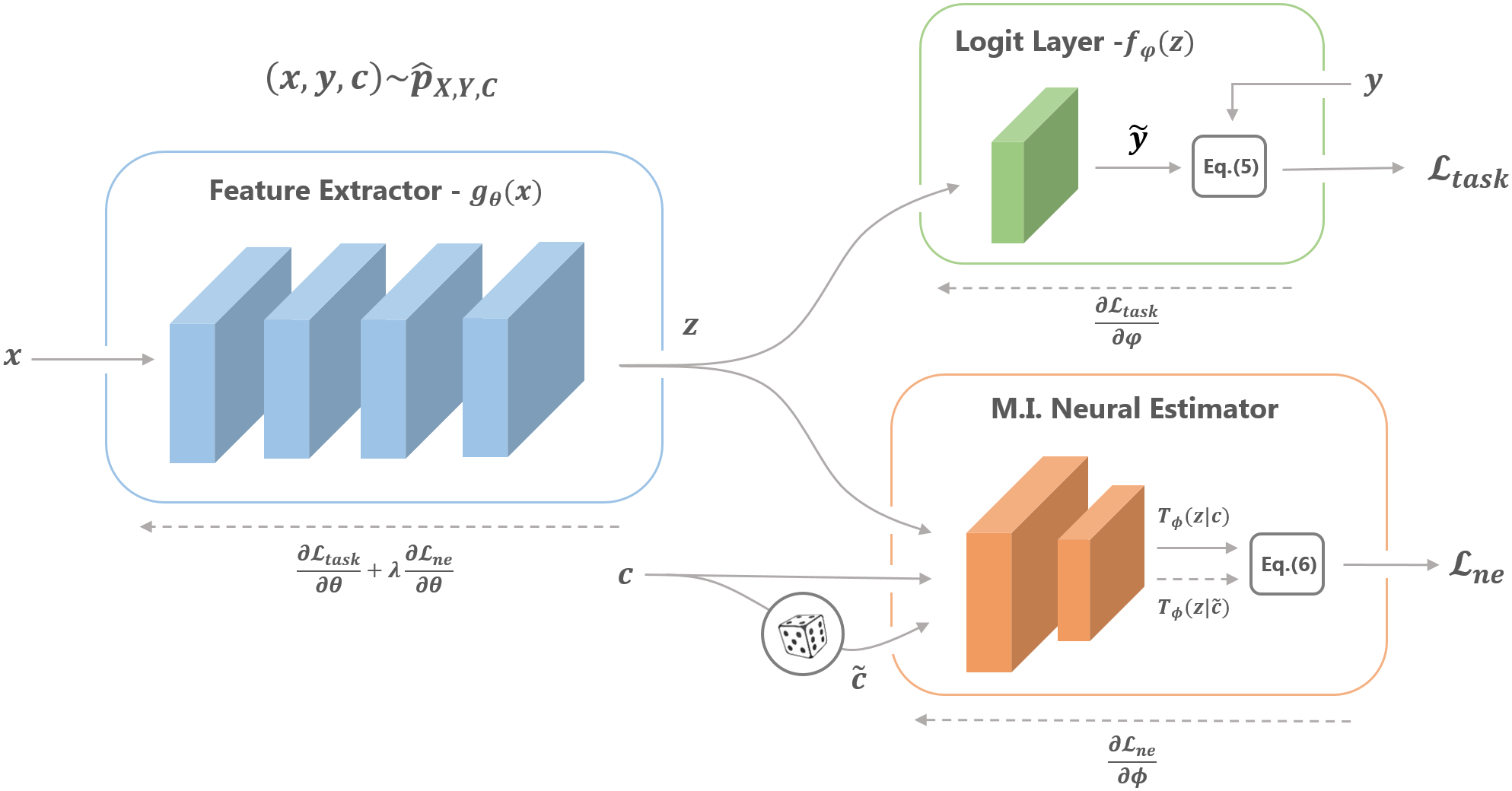}
    \caption{ \footnotesize \textbf{Model overview.} The neural network devised for the given task is the concatenation of the blue module (feature extractor $g_{\theta}$) and the green module (logit layer $f_{\psi}$). Solid lines indicate the forward flow, dashed lines indicate gradient backpropagations. The feature extractor takes in input samples $\mathbf{x}$ and outputs feature vectors $\mathbf{z}$. The logit layer takes in input the feature vectors and outputs predictions $\mathbf{\tilde{y}}$. To optimize for the given task, these modules can be trained by minimizing the cross-entropy between predictions and labels $\mathbf{y}$. The orange module~\cite{belghazi18a} estimates the mutual information between the feature vectors $\mathbf{z}$ and the attributes $\mathbf{c}$. To estimate the mutual information, $T_{\phi}$ processes the concatenation of feature vectors and attributes from the joint distribution and the marginals. Following Belghazi et al.~\cite{belghazi18a}, we approximate sampling from the marginal by shuffling the batch of attributes ($\mathbf{\tilde{c}}$). The estimation of the mutual information is the maximum w.r.t. $\phi$ of the output of the orange module $\mathcal{L}_{ne}$.}\vspace{20pt}
    \label{fig:model}
\end{figure*}

In order to proceed with a more tractable problem, we consider the Lagrangian of Eq.~\eqref{eq:obj} 
\begin{equation}\label{eq:lagr}
\min_{\theta,\psi} \mathcal{L} \coloneqq \mathcal{L}_{task}(\theta,\psi) + \lambda I(Z,C)
\end{equation}
where the first term is a loss associated with the task of interest, whose minimization ensures that the learned representation is sufficient for our purposes. The second term is the mutual information between the learned representation and the given attributes. The hyper-parameter $\lambda$ balances the trade-off between optimizing for a given task and minimizing the mutual information. 

Concerning the first term of the objective, we will consider classification tasks throughout this work, and thus we assume that our aim is minimizing the cross-entropy loss between the output of the model $\mathbf{\tilde{y}}$ and the ground truth $\mathbf{y}$. 
\begin{equation}\label{eq:task}
\mathcal{L}_{task} \coloneqq \frac{1}{N} \sum^{N}_{i=1}  \mathbf{y}_i^T \log s(\mathbf{\tilde{y}}_i)
\end{equation}
where $s$ is the softmax function and $N$ is the number of given datapoints.

Concerning the second term of the objective in Eq.~\eqref{eq:obj}, as already mentioned, the analytical formulation of the mutual information is of scarce utility to evaluate $I(Z,C)$. Indeed, we do not explicitly know the probability distributions that the learned representation and the attributes obey to. Therefore, we need an estimator for the mutual information $\widehat{I}(Z,C)$, with the requirement of being differentiable with respect to the model parameters $\theta$. 

In order to attain our targeted goal, we take advantage of the work by Belghazi et al.~\cite{belghazi18a} (Eq.~\eqref{eq:MINE-approx}), and exploit a second neural network $T_{\phi}$ (``statistics network'') to estimate the mutual information. We therefore introduce an additional loss function
\begin{align}
\mathcal{L}_{ne} &\coloneqq \mathbb{E}_{(\mathbf{z},\mathbf{c}) \sim \hat{p}_{Z,C}} [T_{\phi}(\mathbf{z}|\mathbf{c})] - \log \mathbb{E}_{\mathbf{z}\sim\hat{p}_Z, \mathbf{\tilde{c}}\sim\hat{p}_C} [\exp T_{\phi}(\mathbf{z}|\mathbf{\tilde{c}})]\label{eq:MINE_loss}
\end{align}
that, once maximized, provides an estimate of the mutual information
\begin{equation}\label{eq:max}
\widehat{I}_{ne}(Z,C) = \max_\phi \mathcal{L}_{ne}.
\end{equation}
In Eq.~\eqref{eq:MINE_loss}, the notation $\hat{p}$ reflects that we rely on the empirical distributions of features and attributes, the operator ``$|$'' indicates vector concatenations and ``ne'' stands for ``neural estimator''~\cite{belghazi18a}. The loss $\mathcal{L}_{ne}$ also depends on $\theta$, since Eq.~\eqref{eq:MINE_loss} depends on $\mathbf{z}$. Combining the pieces together, we obtain the following problem 
\begin{align}
\min_{\theta,\psi} \{\mathcal{L}_{task}(\theta,\psi) + \lambda \widehat{I}_{ne}(Z,C)\} = \underbrace{\min_{\theta,\psi} \{ \mathcal{L}_{task}(\theta,\psi)   + \lambda \underbrace{\max_\phi \mathcal{L}_{ne}(\phi, \theta)}_{\text{MI estimation}}}_{\text{Representation learning}}\} \label{eq:final_obj}
\end{align}
Intuitively, the inner maximization problem ensures a reliable estimate of the mutual information between the learned representation and the attributes. The outer minimization problem is aimed at learning a representation that is at the same time optimal for the given task and unbiased with respect to the attributes.   

\vspace{-10pt}
\subsection{Implementation Details}\label{sez:impl-det}


Concerning the modules introduced in Section~\ref{sec:prob_form}, we implement the feature extractor $g_\theta$ (which computes features $\mathbf{z}$ from datapoints $\mathbf{x}$) and the classifier $f_\psi$ (which predicts labels $\tilde{\mathbf{y}}$ from $\mathbf{z}$) as feed-forward neural networks. The classifier $f_\psi$ is implemented as a shallow logit layer to accomplish predictions on the task of interest. As already mentioned, the model $T_{\phi}$ is also a neural network; it accepts in input the concatenation of feature vectors $\mathbf{z}$ and attribute vectors $\mathbf{c}$, and through Eq.~\eqref{eq:MINE_loss} allows estimating the mutual information between the two random variables. The nature of the modules allow to optimize the objective functions in~\eqref{eq:final_obj} via backpropagation~\cite{BackProp}. Figure~\ref{fig:model} portrays the connections between the different elements, and how the losses~\eqref{eq:task} and~\eqref{eq:MINE_loss} originate.

A crucial point that needs to be addressed when jointly optimizing the two terms of Eq.~\eqref{eq:final_obj} is that, while the distribution of the attributes $\widehat{p}_C$ is static, the distribution of the feature embeddings $\widehat{p}_Z$ depends on $\theta$, which changes throughout the learning trajectory. For this reason, the mutual information estimator needs to be constantly updated during training, because an estimate $\widehat{I}_{ne}(Z_t,C)$, associated with $\theta_t$ at step $t$, is no longer reliable at step $t+1$. To cope with this issue, we devise an iterative procedure where, prior to every gradient descent update on $(\theta, \psi)$, we update MINE on the current model, through the inner maximizer in Eq.~\eqref{eq:final_obj}. This guarantees a reliable mutual information estimation. 
\begin{algorithm}[t]
\caption{Learning Unbiased Representations}
\label{alg:train_proc}
\begin{spacing}{1.0}
\begin{algorithmic}[1]
\small
\State \textbf{Input:} Dataset $\{(\mathbf{x}^{(i)},\mathbf{y}^{(i)},\mathbf{c}^{(i)})\}_{i=1}^{N}$, initialized weights $\theta_0$, $\psi_0$, $\phi_0$, learning rates $\alpha$, $\eta$, hyper-parameters $\lambda,K,T$.
\State \textbf{Output:} learned weights $\theta$, $\psi$
\State \textbf{Initialize:} $\theta \gets \theta_0$, $\psi \gets \psi_0$, $\phi \gets \phi_0$
\For{$\text{t}=1,...,T$}

    \For{$\text{k}=1,...,K$} (train MINE) 

        \State sample mini-batches $\{(\mathbf{x}^{(i)},\mathbf{c}^{(i)})\}_{i=1}^{m}$, $\{\mathbf{\tilde{c}}^{(i)}\}_{i=1}^{m}$

        \State evaluate $\mathcal{L}_{ne}$ (Eq.~\eqref{eq:MINE_loss})

        \State $\phi \gets \phi + \eta \nabla_{\phi} \mathcal{L}_{ne}$

    \EndFor

    \State sample mini-batches $\{(\mathbf{x}^{(i)},\mathbf{x}^{(i)},\mathbf{c}^{(i)})\}_{i=1}^{n}$, $\{\mathbf{\tilde{c}}^{(i)}\}_{i=1}^{n}$

    \State evaluate $\mathcal{L}_{task}$ (Eq.~\eqref{eq:task}) and $\mathcal{L}_{ne}$ (Eq.~\eqref{eq:MINE_loss})
    \State $\theta \gets \theta - \alpha\nabla_{\theta}(\mathcal{L}_{task} + \lambda \mathcal{L}_{ne})$ 
    \State $\psi \gets \psi - \alpha\nabla_{\psi}\mathcal{L}_{task}$

\EndFor


\end{algorithmic}
\end{spacing}
\end{algorithm}

As already mentioned, one key difference with adversarial methods is that we can train MINE until convergence prior to each gradient descent step on the feature extractor, without the risk of obtaining gradients whose magnitude is close to zero~\cite{arjovsky2017iclr}, since our estimator is not a discriminator (being the mutual information unbounded, sometimes gradient clipping is actually beneficial~\cite{belghazi18a}). The full training procedure is detailed in Algorithm~\ref{alg:train_proc}.\\

\noindent
\textbf{Training techniques.} Before discussing our results, we briefly comment below some techniques that we could appreciate to generally increase the stability of the proposed training procedure. While code and hyper-parameters can be found in the Supplementary Material, we believe that the reader can benefit from the discussion.\\~\\
\textit{\textbf{(a)}} Despite MINE~\cite{belghazi18a} can estimate the mutual information between continuous random variables, we observed that the estimation is eased (in terms of speed and stability) if the attribute labels $\mathbf{c}$ are discrete. \textit{\textbf{(b)}} We observed an increased stability in training MINE~\cite{belghazi18a} for lower-dimensional representations $\mathbf{z}$ and attributes $\mathbf{c}$. For this reason, as we will discuss in Section~\ref{sec:exp}, feature extractors with low-dimensional embedding layer are favored in our settings. \textit{\textbf{(c)}} The feature extractor $g$ receives gradients related to both $\mathcal{L}_{task}$ and $\mathcal{L}_{ne}$: since the mutual information is unbounded, the latter may dominate the former. Following Belghazi et al.~\cite{belghazi18a}, we overcome this issue via gradient clipping (we refer to original work for details). \textit{\textbf{(d)}} We observed that training MINE requires large mini-batches: when this was unfeasible due to memory issues, we relied on gradient accumulation. \textit{\textbf{(e)}} We observed that using vanilla gradient descent over Adam optimizer~\cite{AdamOptimizer} eases training MINE~\cite{belghazi18a} in most of our experiments.


\vspace{-5pt}
\section{Experiments}\label{sec:exp}
\vspace{-5pt}

\begin{wrapfigure}[11]{r}{0.5\textwidth}
\vspace{-30pt}
    \centering
    \includegraphics[width=\linewidth]{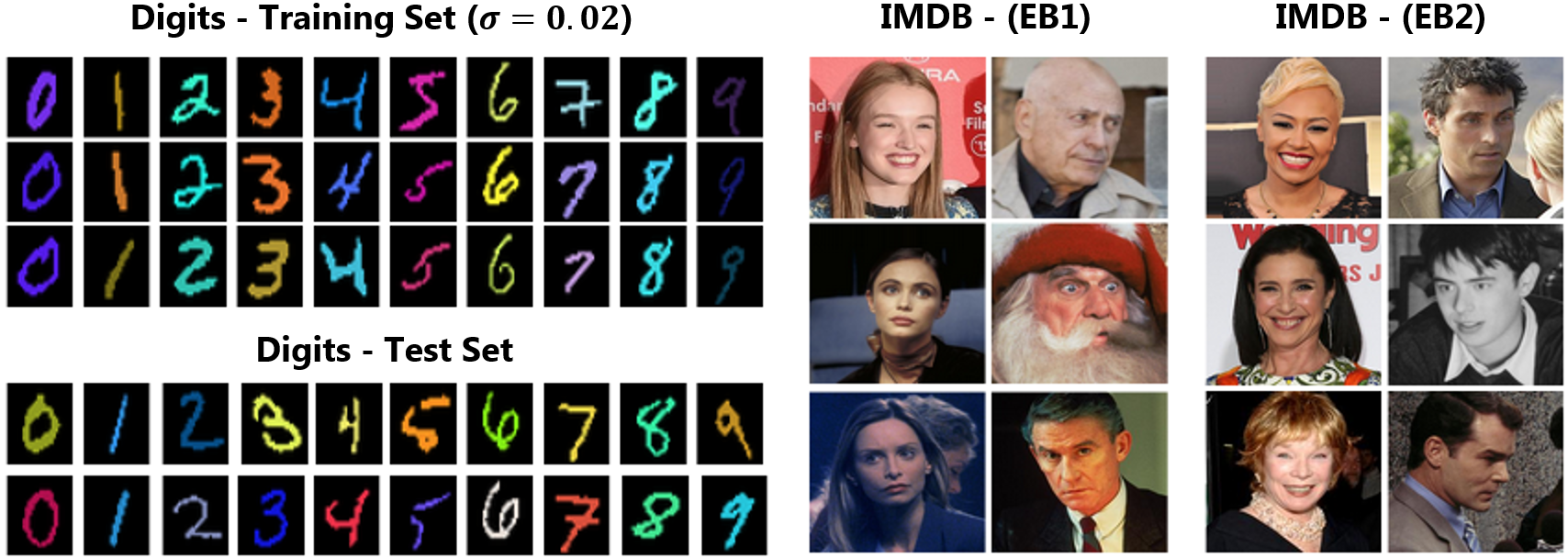}
  \caption{\footnotesize \textit{Left:} digit examples for each class from training (here with $\sigma=0.02$) and test set. \textit{Right:} Women and Men images from the two splits of the training set of the IMDB dataset.}\label{fig:dataset}
\end{wrapfigure}

In the following, we show the effectiveness of models trained via Algorithm~\ref{alg:train_proc} in a series of benchmarks. First, we report results related to the setup proposed by Kim et al.~\cite{kim2019cvpr} -- learning to recognize color-biased digits without relying on color information. Next, we show that our proposed solution can scale to higher-capacity models and more difficult tasks, through the IMBD benchmark~\cite{zisserman2018,kim2019cvpr}, where the goal is classifying people age from images of their face, without relying on the gender bias. Finally, we show that our method can also be applied as it is to learn ``fair'' classifiers, by training models on the German dataset~\cite{german-dataset}.


\vspace{-5pt}
\subsection{Digit Recognition}\label{sec:digit}

\textbf{Experimental setup.} Following the setting defined by Kim et al.~\cite{kim2019cvpr}, we consider a digit classification task where each digit, originally from MNIST \cite{MNIST}, shows an artificially induced color bias. More specifically, in the training set (with $60,000$ samples), digit colors are drawn from Gaussian distributions, whose mean values are different for each class. In the test set (with $10,000$ samples), digits show random colors. The benchmark is designed with seven different standard deviation values $\sigma$ (equally spaced between $0.02$ and $0.05$): the lower the value, the more difficult the task, since the model can fit the training set by recognizing colors instead of shapes, thus poorly generalizing (see Figure~\ref{fig:dataset}). To extract the color information (the attribute $\mathbf{c}$, recalling notation from Section~\ref{sec:prob_form}), the maximum pixel value is encoded in a binary vector with 24-bit (8 bits per channel). Since the background is always black, the maximum value reflects the digit color. 

Concerning the model, we exploit a convolutional neural network~\cite{LeNet} with architecture \textit{conv-pool-conv-pool-fc-fc-softmax}. The output of the second fully connected layer ($\mathbf{z}$) is given in input to both the logit layer and MINE (Figure~\ref{fig:model}). The architecture of the statistics network $T_{\phi}$ in MINE is a multi-layer perceptron (MLP) with 3 layers. More architectural details can be found in the Supplementary Material. We compare models trained via Algorithm~\ref{alg:train_proc} with the solutions proposed by Kim et al.~\cite{kim2019cvpr} and Alvi et al.~\cite{zisserman2018}, averaging across $3$ runs and using accuracy as a metric. Before comparing against related work, we discuss how crucial hyper-parameters can be selected in our setting.\\ 


\noindent
\textbf{Hyper-parameter choice.}
We discuss in the following the model behavior as we modify $\lambda$, that governs the trade-off between learning a task and minimizing the mutual information between features and attributes.
\begin{figure*}[t!]
    \centering
    \includegraphics[width=0.9\textwidth]{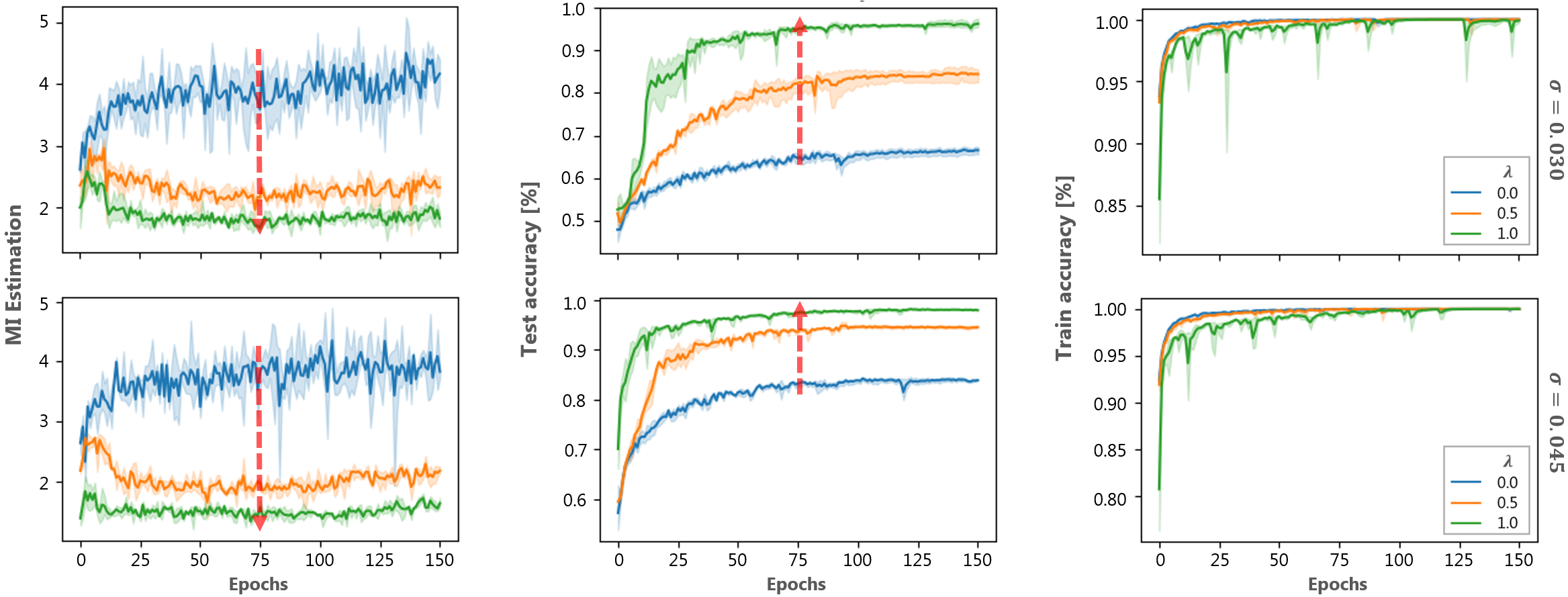}
    \caption{\footnotesize \textbf{Digit experiment -- ablation study.} Evolution of mutual information estimation (left), test accuracy (middle) and training accuracy (right) for models trained on digits with $\sigma=0.03$ and $\sigma=0.045$ (top and bottom, respectively). Models are trained with Algorithm~\ref{alg:train_proc} with $\lambda=0.0$ (baseline, blue), $\lambda=0.5$ (orange) and $\lambda=1.0$ (green). Increasing the value of the hyper-parameter $\lambda$ allows reducing the mutual information between the learned representation ($Z$) and the attributes ($C$). In turn, models better generalize to unbiased samples (test set). Further plots 
    in the Supplementary.}
    \label{fig:digit_ablation}
\end{figure*}
Figure~\ref{fig:digit_ablation} reports the evolution of mutual information estimation (left), accuracy on test samples (middle) and accuracy on training sample (right) for models trained with $\lambda=0.0,0.5,1.0$ in blue, orange and green, respectively, for $\sigma=0.03,0.045$ (top and bottom, respectively). It can be observed that the mutual information between embeddings $\mathbf{z}$ and color attributes $\mathbf{c}$ can be reduced by increasing $\lambda$. Importantly, this results in a significantly higher accuracy on (unbiased) test samples. The importance of this result is twofold: on the one hand, it is a proof of concept of the intuition that lowering the mutual information does help generalizing to unbiased sources; on the other, it provides us with a possible cross-validation strategy to pick a proper $\lambda$ value (the one that allows minimizing the mutual information more efficiently). As can be observed in the plots on the right, the training procedure becomes more unstable when we increase $\lambda$. Therefore, in order to select the proper hyper-parameter, we can choose the highest $\lambda$ value that allows the model fitting the data (\textit{i.e.}, minimizing $\mathcal{L}_{task}$) and reducing the mutual information (\textit{i.e.}, minimizing $\mathcal{L}_{ne}$).

Another important hyper-parameter is the number of iterations used to train MINE~\cite{belghazi18a} prior to each gradient update on the feature extractor ($K$ in Algorithm~\ref{alg:train_proc}). We observed that, in general, the higher the number of iterations the better. This was expected, because the estimate of the mutual information becomes more reliable. In the results proposed in the following paragraph, we set $K=80$. We refer to the Supplementary Material for details regarding the other less critical hyper-parameters.


\renewcommand{\arraystretch}{1.3}
\begin{table*}[t]
\footnotesize
\begin{center}
\caption{\footnotesize \textbf{Digit experiment -- comparison with related work.} Experimental results on colored digit classification for different levels of variance ($\sigma$) in the color distribution. The first row reports results  related to models trained via standard Empirical Risk Minimization (ERM). The second row reports results obtained by Alvi et al.~\cite{zisserman2018}. The third row reports the results published by Kim et al.~\cite{kim2019cvpr}. The last row reports results achieved with our method (with $\lambda=1.0$.)} 
\label{tab:digit_results}
\vspace{-10pt}
\resizebox{\textwidth}{!}{
\begin{tabular}{rcccccccccccccc}

\multicolumn{8}{c}{} \\

\toprule

& \multicolumn{7}{c}{\textbf{Color variance}} \\
\cmidrule(r){2-8}

\textbf{Training} & $\mathbf{\sigma=0.020}$ & $\mathbf{\sigma=0.025}$ & $\mathbf{\sigma=0.030}$ & $\mathbf{\sigma=0.035}$ & $\mathbf{\sigma=0.040}$  & $\mathbf{\sigma=0.045}$  & $\mathbf{\sigma=0.050}$ \\
\midrule
ERM ($\lambda=0.0$) & 0.476 $\pm$ 0.005 & 0.542 $\pm$ 0.004 & 0.664 $\pm$ 0.007 & 0.720 $\pm$ 0.010 & 0.785 $\pm$ 0.003 & 0.838 $\pm$ 0.002 & 0.870 $\pm$ 0.001\\

Alvi et al.~\cite{zisserman2018} & 0.676 & 0.713 & 0.794 & 0.825 & 0.868 & 0.89 & 0.917 & \\

Kim et al.~\cite{kim2019cvpr}& 0.818 & 0.882 & 0.911 & 0.929 & 0.936 & 0.954 & 0.955 & \\

\textit{Ours} & $0.864 \pm 0.052$ & $0.925 \pm 0.020$ & $0.959 \pm 0.008$ & $0.973 \pm 0.003$ & $0.975 \pm 0.001$ & $0.980 \pm 0.001$ & $0.982 \pm 0.001$\\

\bottomrule
\end{tabular}}
\end{center}
\end{table*}

\vspace{2pt}
\noindent
\textbf{Comparison with related work.} We report in Table~\ref{tab:digit_results} the comparison between our method with $\lambda=1.0$ and related works~\cite{kim2019cvpr,zisserman2018}. We can observe consistently improved results in all the benchmarks (different $\sigma$'s). We emphasize that our method is more effective as the bias is more severe (small $\sigma$'s). It is also important to stress that Kim et al.~\cite{kim2019cvpr} do not introduce any strategy to search the hyper-parameters that balance the adversarial game, whereas in this work the hyper-parameter search is efficiently resolved. Furthermore, the authors do not report any statistics around their results (\eg, average and standard deviation across different runs), making a fair comparison difficult.  




\vspace{-10pt}
\subsection{IMDB: Removing the Age Bias}

\textbf{Experimental setup.} Following related works~\cite{zisserman2018,kim2019cvpr}, we consider the IMDB dataset~\cite{imdb_dataset} as benchmark. It contains cropped images of celebrity faces with ground truth annotations related to gender and age. Alvi et al.~\cite{zisserman2018} consider two subsets of the training set that are severely biased for what concerns age: the EB1 (``Extreme Bias'') split ($36,003$ samples) only contains images of women with an age in the range 0-30, and men who are older than 40; vice versa, the EB2 split ($16,799$ samples) only contains images of men with an age in the range 0-30, and women older than 40 (see Figure \ref{fig:dataset}). The test set ($22,468$ samples) contains faces without any restrictions on age/gender (uniformly samples). The goal here is learning an age-agnostic model, to overcome the bias present in the dataset.

Following previous work~\cite{zisserman2018,kim2019cvpr}, we encode the age attribute (our biased attribute, $\mathbf{c}$) using bins of 5 years, via one-hot encoding. We use a ResNet-50~\cite{ResNet} model pre-trained on ImageNet \cite{ImageNet} as classifier, modified with a 128-dimensional fully connected layer before the logit layer. This narrower embedding serves as our $\mathbf{z}$, and the reduced dimension eases the estimation of the mutual information, while not causing any detrimental effect in terms of accuracy. 
For each split (EB1 and EB2), we train the model through Algorithm~\ref{alg:train_proc} and evaluate it on the test set and on the split not used for training. We followed the same procedure detailed in Section~\ref{sec:digit} to choose the hyper-parameter $\lambda$, obtaining $\lambda=0.5$ and $\lambda=0.9$ for EB1 and EB2 splits, respectively; we set $K=40$. We compare our results with the ones published by related works~\cite{zisserman2018,kim2019cvpr}, using accuracy as a metric. We limited the training sets to only $2,000$ samples: this choice was due to the fact that with the whole training sets we could observe baselines ($\lambda=0.0$) significantly higher than published results~\cite{kim2019cvpr}, whereas they are comparable for models trained on a subset.\\

\begin{figure}[t!]
\captionsetup{type=table}
\caption{\footnotesize \textbf{IMDB experiment.} \emph{(Table on the Left).} We compare against related work in the Table on the left. The first row reports results obtained by setting $\lambda=0.0$ (ERM baseline). The last row reports results obtained with our method, 
by setting $\lambda=0.5$ for EB1 and $\lambda=0.9$ for EB2. Each column reports results associated with the indicated test set. \emph{(Plots on the Right)}. Train on EB2. \emph{(Bottom-Right).} Evolution over iterations of the test accuracy reported on the last column for our method ($\lambda=0.9$, green) and baseline models ($\lambda=0.0$, blue). \emph{(Top-Right).} The mutual information is closer to $0$ when using our method. Our results were averaged over $3$ different runs.}
\vspace{-5pt}
\begin{minipage}{0.0001\linewidth}
~
\end{minipage}
  \begin{minipage}{0.60\linewidth}
    \centering
        \scriptsize
        \renewcommand{\arraystretch}{1.4}
            \begin{tabular}{rcccc}
            \toprule
            & \multicolumn{2}{c}{\textbf{Train on EB1}} & \multicolumn{2}{c}{\textbf{Train on EB2}} \\
            \cmidrule(r){2-3} \cmidrule(r){4-5}
            \textbf{Method} & \textbf{EB2} & \textbf{Test} & \textbf{EB1} & \textbf{Test} \\
            \midrule
            
            ERM ($\lambda=0.0$) & $0.650\pm0.020$ & $0.849\pm0.007 $ & $0.576\pm0.013$ & $0.708\pm0.008$ \\
            
            Alvi et al.~\cite{zisserman2018} & $0.637$~\cite{kim2019cvpr}& $0.856$~\cite{kim2019cvpr} & $0.573$~\cite{kim2019cvpr} & $0.699$~\cite{kim2019cvpr} \\
            
            Kim et al.~\cite{kim2019cvpr} & $0.680 $ & $0.867 $ & $0.642$ & $0.745$ \\
            
            \textit{Ours} & $0.684 \pm 0.010$ & $0.872 \pm 0.010$ & $0.631 \pm 0.036$ & $0.745 \pm 0.022$ \\
            
            \bottomrule
            \end{tabular}
  \end{minipage}
  \begin{minipage}{0.14\linewidth}
  \hspace{1cm}
  \end{minipage}
  \begin{minipage}{0.22\linewidth}
    \includegraphics[width=\linewidth]{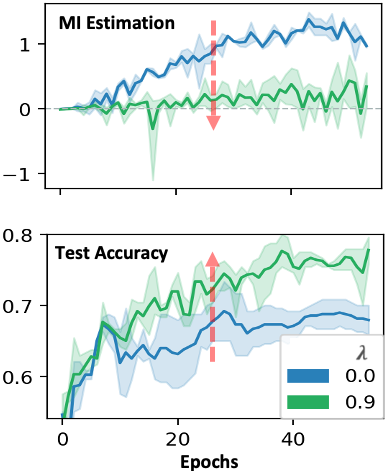}
  \end{minipage}
\vspace{10pt}
\label{tab:imdb}
\end{figure}

\noindent
\textbf{Results.} Table~\ref{tab:imdb} reports our results. In all our experiments, we observe accuracy improvements with respect to the baseline ($\lambda=0.0$). In general, training on one split and testing on the other is more challenging than testing on the (neutral) test set, as confirmed by the baseline results (ERM, first row). In all the different protocols, our method (last row) has superior performance than Alvi et al.~\cite{zisserman2018}, and comparable performance with Kim et al.~\cite{kim2019cvpr}. 

These results confirm that our method can effectively remove biased, detrimental information even when modeling more complex data with higher-capacity models. In this case though, the improvements are more limited than the ones we showed in the digit experiment. One of the reasons might be that age and gender information cannot be decoupled as efficiently as shape and color. In other words, removing age information may not always bring accuracy improvements.
%


\vspace{-20pt}
\subsection{Learning Fair Representations}

\textbf{Experimental setup.} We explored the potentiality of our method in the context of algorithmic fairness with the popular UCI dataset German \cite{german-dataset}. The dataset is composed of $1,000$ samples of customer descriptions with both categorical and continuous attributes. The binary, ground truth label is the risk degree associated with a customer, either good or bad. The goal is to learn a model to predict the customer rate with the constraint of removing the information about the customer age (binarized according to $\gtrless25$). This problem is different with respect to the previous ones: here the invariance towards sensitive attribute does not imply a better generalization on the test set as it happens with, \eg, digit recognition. The removal of the protected attribute is done for the sake of obtaining a fair representation~\cite{kleinberg2016inherent,donini2018empirical,zhang_2018_fairness,wang_2019_fairness}. 

\begin{figure}[!t]
\captionsetup{type=table}
\caption{\footnotesize \textbf{Fairness experiment -- comparison with related work and ablation study.} 
\emph{(Table on the Left).} We compare against results as reported in~\cite{mary2019fairness}. For accuracy (first row), the higher the better. For EO (last row), the lower the better (\textit{i.e.}, the ``fairer''). \emph{(Plots on the Right).} The barplots show how the two considered metrics vary as we modify the hyper-parameter $\lambda$. EO (top) is significantly reduced as we set higher values of $\lambda$. Vice versa, test accuracy (bottom) is only slightly affected. Our results were averaged across $10$ different runs.}
\vspace{-10pt}
\begin{minipage}{0.025\linewidth}
~
\end{minipage}
  \begin{minipage}{0.60\linewidth}
    \centering
        \scriptsize
        \begin{tabular}{rccccc}
        \multicolumn{6}{c}{\footnotesize{\textbf{German experiment}}} \\
        \toprule
        & \multicolumn{5}{c}{\textbf{Method}} \\
        \cmidrule(r){2-6}
        & SVM~\cite{donini2018empirical} & FERM~\cite{donini2018empirical} & NN~\cite{mary2019fairness} & NN + $\chi^2$~\cite{mary2019fairness} & \textit{Ours ($\lambda=1.0$)}\\
        \midrule
        Acc. & $0.74 \pm 0.03$ & $0.73 \pm 0.04$ & $0.74 \pm 0.04$ & $0.73 \pm 0.03$ & $0.72\pm 0.03$ \\
        EO & $0.10 \pm 0.06 $ & $0.05 \pm 0.03$ & $0.47 \pm 0.19$ & $0.25 \pm 0.14$ & $0.05\pm 0.05$ \\
        \bottomrule
        \end{tabular}
  \end{minipage}
  \begin{minipage}{0.12\linewidth}
  \hspace{1cm}
  \end{minipage}
  \begin{minipage}{0.18\linewidth}
    \includegraphics[width=\linewidth]{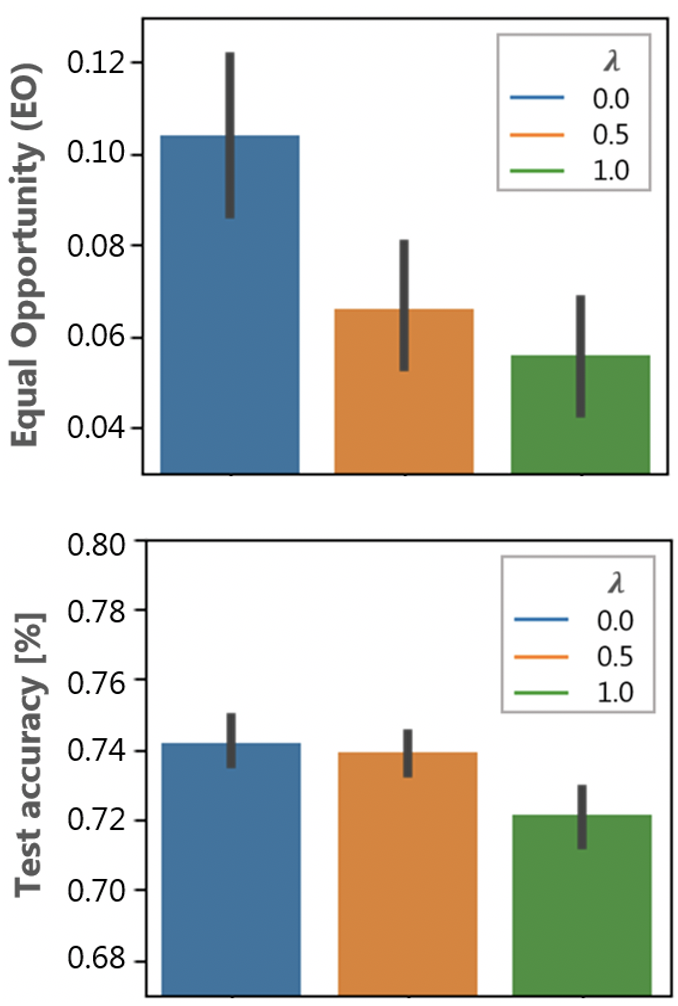}
  \end{minipage}
\vspace{10pt}
\label{tab:german}
\end{figure}

Following previous work, we implemented the feature extractor as a single-layer MLP with 64 units in the hidden layer. MINE's statistics network is a shallow network with 64 hidden units. We randomly split the dataset in 70\% training samples and 30\% test samples, and use accuracy and Equal Opportunity (EO)\footnote{Equal Opportunity measures the discrepancy between the TP rates of ``protected'' and ``non-protected'' populations. Here, $\text{EO}=|\text{TP}(\text{young}) - \text{TP}(\text{not young})|$.} as comparison metrics, averaging across $10$ different runs. The goal is to find a balance between reducing EO (\textit{i.e.}, learning a fairer representation) without observing a too severe decrease in accuracy.\\

\noindent
\textbf{Results.} In the right plots of Table~\ref{tab:german}, we show how the performance varies when increasing $\lambda$ from $0$ (standard Empirical Risk Minimization) to $1$. It can be observed that our method allows training fairer models (\ie, reduced EO), while maintaining a good performance on test. For $\lambda=0.5$, the fairness price is close to zero (\ie, the accuracy does not decrease), while the fairness is substantially improved. We report the comparison with related works on Table~\ref{tab:german} (left). Notice that the FERM method~\cite{donini2018empirical} directly optimizes for fairness, while we do not. This experiment is a proof of concept to show that the fairness community might benefit from our approach, although our main goal is bias removal in contexts where it can improve the model's generalization capabilities.  


\vspace{-10pt}
\section{Conclusions}\label{sec:conclusion}
\vspace{-5pt}

We propose a training procedure to learn representations that are not biased towards dataset-specific attributes. We leverage a neural estimator for the mutual information~\cite{belghazi18a}, devising a method that can be easily implemented in arbitrary architectures, and that relies on a training procedure which is more principled and reliable than adversarial training. When compared with the state of the art~\cite{zisserman2018,kim2019cvpr}, it shows competitive results, with the advantage of a robust hyper-parameter selection procedure. Moreover, the proposed solution has competitive performance even in the fairness setting, where the goal is to find a trade-off between attribute invariance and accuracy.



%
%
\bibliographystyle{splncs04}
\bibliography{egbib}

\setlength{\textfloatsep}{8pt}

\title{Supplementary Material}
\author{}
\institute{}
\maketitle

\section{Implementation Details}

\begin{figure}[h!]\label{fig:fig1}
    \centering
    \begin{overpic}[width=\textwidth]{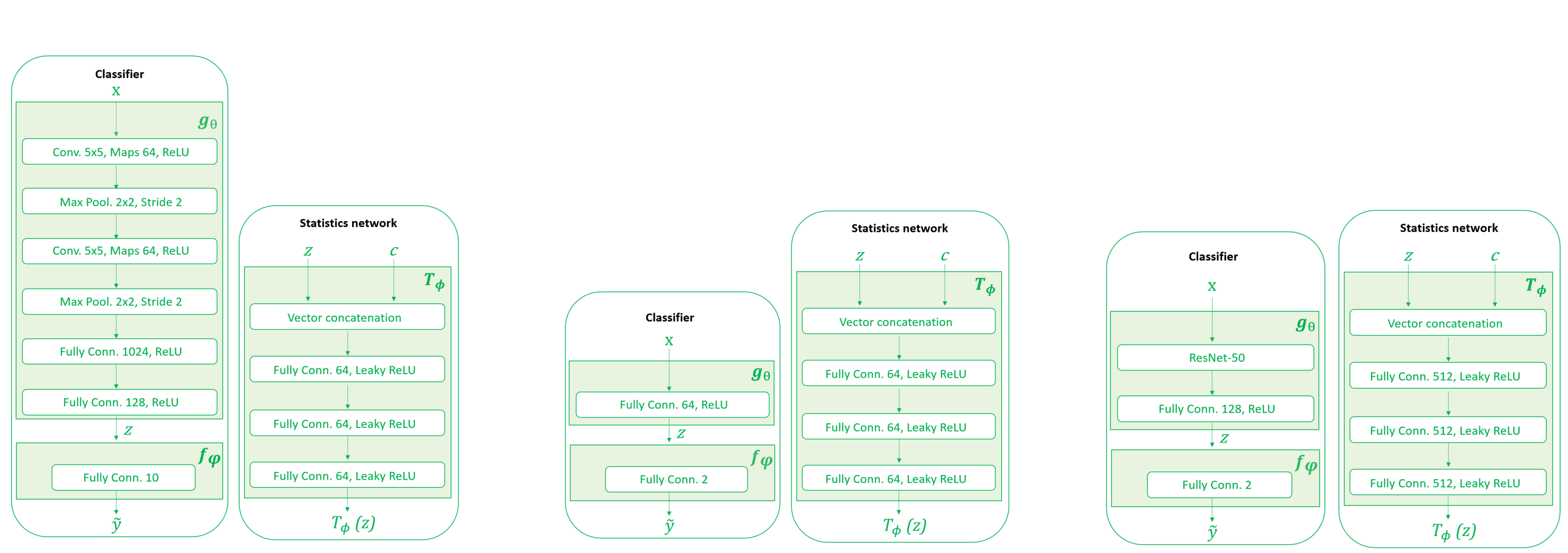}
    \put(13,-2){Digit}
    \put(47.8,-2){IMDB}
    \put(81.5,-2){German}
    \end{overpic}\\~~~~~~~\\
    \caption{\footnotesize Description of the architectures (classifiers and statistics networks) for the experiments on Digits \textit{(left)}, IMDB \textit{(middle)} and German \textit{(right)}.}
    \label{fig:arch}
\end{figure}

In the following paragraphs, we provide the implementation details. We carried out all of our experiments using TensorFlow~\footnote{https://www.tensorflow.org/}. Concerning the architectures used, please refer to Figure \ref{fig:arch}. 

To ease the discussion, we can divide the optimization problem presented in our work into the following two
\begin{equation}\label{o1}
\min_{\theta,\psi} \mathcal{L}_{task} + \lambda \mathcal{L}_{ne}
\end{equation}
\begin{equation}\label{o2}
\max_\phi \mathcal{L}_{ne}
\end{equation}
where the learning rates associated to \eqref{o1} and \eqref{o2} are $\alpha$ and $\eta$, respectively. We use the same notation of Algorithm 1 (in the paper).\\

\noindent
\textbf{Digit experiment.} We train our models for 150 epochs, using mini-batches of size $1024$. The learning rates $\alpha$ and $\eta$ are both set to $10^{-4}$. We use Adam~\cite{AdamOptimizer} as optimizer for \eqref{o1} and \eqref{o2}. For each gradient update to optimize~\eqref{o1} with respect to $\theta,\psi$, we update MINE parameters $80$ times ($K=80$). That is, we perform 80 update steps to optimize~\eqref{o2}, as to better train MINE (see Section \ref{sez:hyper} for a detailed discussion around this choice).\\

\noindent
\textbf{IMDB experiment.} For both training splits (EB1 and EB2) we restrict the training set to $2000$ samples 
This choice is motivated by the fact that using the whole training sets we observed higher baselines results then the ones published in previous art~\cite{kim2019cvpr}. We trained each model for $6$ epochs with mini-batch size set to $24$. The learning rate $\alpha$ is set to $10^{-5}$; the learning rate $\eta$ is set to $10^{-1}$. We use Adam~\cite{AdamOptimizer} as optimizer for~\eqref{o1} and vanilla gradient descent for~\eqref{o2}. We found a number $K=20$ of MINE iterations to be sufficient in order to estimate the mutual information throughout training.\\

\noindent
\textbf{German experiment.} We adopted the same settings as previous art that uses this benchmark~\cite{moyer2018neurips}. The $1,000$ data samples available are split in $70\%$ training and $30\%$ test (randomly picked in each run). The model is trained for $500$ epochs with mini-batch size set to $64$. The learning rate $\alpha$ is set to $10^{-5}$; the learning rate $\eta$ is set to $10^{-1}$. We use Adam~\cite{AdamOptimizer} as optimizer for \eqref{o1}, and vanilla gradient descent for~\eqref{o2}. We set to a number of MINE iterations $K=30$.

\section{Discussion on the Hyper-Parameters}\label{sez:hyper}
\noindent
In this section, we discuss the hyper-parameters that we adopted throughout the experiments reported in this work.\\

\begin{figure}[h!]\label{fig:fig1}
    \centering
    \includegraphics[width=\textwidth]{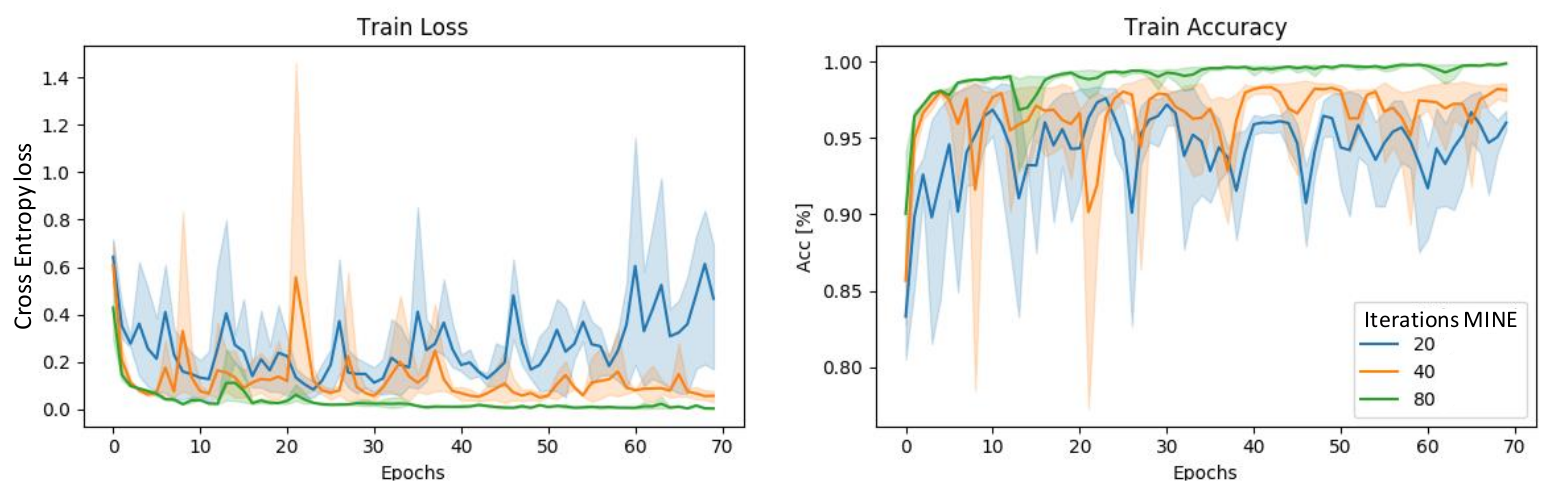}
    \caption{\footnotesize Training (cross-entropy) loss \textit{(left)} and training accuracy \textit{(right)} with $\lambda=1.0$ for different number of iterations of MINE ($K$) on the digit recognition task (setting $\sigma=0.02$). An increased number of iterations ($K=20,40,80$ in \textit{blue}, \textit{orange} and \textit{green}, respectively) has the effect of stabilizing the training procedure, \ie it allows the model minimizing the loss function and fitting the training data. The charts report the average of 3 runs.}
    \label{fig:iter}
\end{figure}

\noindent
\textbf{Choice of the number of iterations to update MINE.} We found that increasing the number of iterations to estimate $I(Z,C)$ stabilizes the overall training procedure, as shown in Figure~\ref{fig:iter}. As our intuition behind this fact, we posit that the better the estimation of the mutual information through MINE is, the more precise and effective the gradients $\nabla_{\theta}\mathcal{L}_{ne}$ are. The only drawback we observed is the increased computational cost, since the time increases linearly with the number of iterations employed to estimate the mutual information.\\

\begin{figure}\label{fig:fig1}
    \centering
    \begin{overpic}[width=\textwidth]{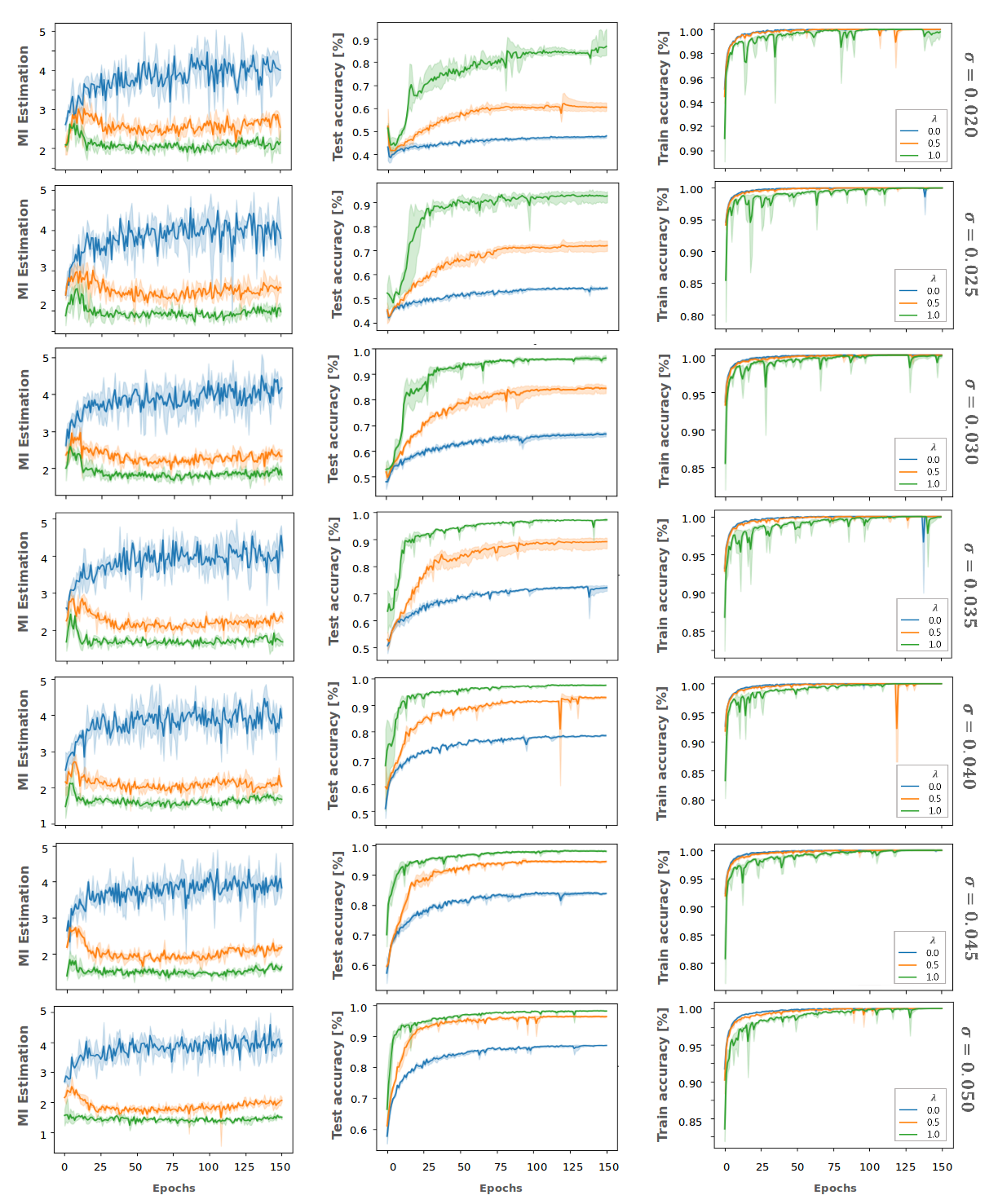} 
    \end{overpic}
    \caption{\footnotesize Values for mutual information (left column), test accuracy (middle column) and train accuracy (right column). We accounted for the different color, modelled by different $\sigma$ (check Section 5 of the paper), and here represented by different rows. It is visible how a decrease in the (estimated) mutual information correlates with an improved performance.}
    \label{fig:lambda_parameter}
\end{figure}

\noindent
\textbf{Choice of the hyper-parameter $\lambda$.}
The hyper-parameter $\lambda$ regulates the trade-off between minimizing the task loss and reducing the mutual information between the biased attribute and the learned representation in~\eqref{o1}. In Section 5 of the paper, we describe how to properly tune it. We report in Figure \ref{fig:lambda_parameter} the complete version of the analysis reported in the manuscript for the Digit experiment. We report the evolution of mutual information, test accuracy and training accuracy for different values of the hyper-parameter $\lambda$, while $\sigma$ is fixed to be equal to one of the following values: $0.020, 0.025, 0.030, 0.035, 0.040$ or $0.45$.

\end{document}